\pdfoutput=1

\documentclass[11pt]{article}

\usepackage[final]{naacl2021}
\usepackage{comment}

\usepackage[utf8]{inputenc} 
\usepackage[T1]{fontenc}    
\usepackage{hyperref}       
\usepackage{url}            
\usepackage{booktabs}       
\usepackage{amsfonts}       
\usepackage{nicefrac}       
\usepackage{microtype}      
\usepackage{graphicx}
\usepackage{subcaption}
\usepackage{verbatim}
\usepackage{times}
\usepackage{latexsym}

\newcommand\nnfootnote[1]{%
  \begin{NoHyper}
  \renewcommand\thefootnote{}\footnote{#1}%
  \addtocounter{footnote}{-1}%
  \end{NoHyper}
}

\title{Multitask Learning for Citation Purpose Classification}

\renewcommand{\thefootnote}{*}

\author{

  Alex Oesterling\textsuperscript{*}, Angikar Ghosal\textsuperscript{*}, Haoyang Yu\textsuperscript{*}, Rui Xin\textsuperscript{*},  Yasa Baig\textsuperscript{*},\\ \textbf{Lesia Semenova, Cynthia Rudin}    \\
  Duke University \\
  \texttt{\{alex.oesterling,angikar.ghosal,haoyang.yu079,rui.xin926,  } \\
  \texttt{yasa.baig,lesia.semenova,cynthia.rudin\}@duke.edu}
}

\begin{document}

\maketitle
\begin{abstract}


We present our entry into the 2021 3C Shared Task \textit{Citation Context Classification based on Purpose} competition.  The goal of the competition is to classify a citation in a scientific article based on its purpose. This task is important because it could potentially lead to more comprehensive ways of summarizing the purpose and uses of scientific articles,
but it is also difficult, mainly due to 
the limited amount of available training data in which the purposes of each citation have been hand-labeled, along with the subjectivity of these labels. Our entry in the competition is a multi-task model that combines multiple modules designed to handle the problem from different perspectives, including hand-generated linguistic features, TF-IDF features, and an LSTM-with-attention model. We also provide an ablation study and feature analysis whose insights could lead to future work.

\end{abstract}

\nnfootnote{\textsuperscript{*}Equal contribution.}

\section{Introduction}

The influence of a scientific article has often been measured by its citation count. Citation counts can be highly impactful, influencing the overall ratings of journals (e.g., their Impact Factor and Eigenfactor) and are used to quantify the productivity and influence of authors (e.g., h-index, i-index).
However, these methods all operate under the false premise that all citations should be counted equally. Treating all citations with equal weight ignores the wide variety of functions that citations perform. Citations can provide information about the methodology used in the work, point to a contrasting perspective, provide motivation or background information on a field, describe a previous work that the present work extends, or suggest that future work should consider a specific direction. Because citations are so influential, there is a clear need for automated methods to judge both the role and degree of importance of citations within articles. 

The 3C Shared Task \textit{Citation Context Classification based on Purpose} competition, organized by Scholarly Document Processing @ NAACL 2021, aims to improve the state-of-the-art in automated citation classification \cite{prideknothSCP2021}. Using the full text of an article, the competition's goal is to identify the purpose of each citation. In this paper, we describe our high-scoring entry into this competition, achieving third place.

The paper is organized as follows. In Section \ref{sec:related}, we discuss related work. We describe the competition task and available data in Section \ref{sec:task}. In Section \ref{sec:methodology}, we present our approach and methodology. Section \ref{sec:results} contains our results and an ablation study. Finally, we close the paper with a discussion about our findings in Section \ref{sec:discussion}. 

\section{Related Work}\label{sec:related}
Past work on automated citation classification can be divided into two main categories: approaches centered around hand-generated linguistic features and approaches that utilize deep learning models.

Works that fall into the first category analyze the citation context or the full text of the cited article in order to design strong linguistic features.
More specifically, Teufel et al. \cite{teufel} focus on semantic-based features, identifying cue words, phrases, verb tense, voice, and so on. Hou et al. \cite{hou} introduce count-based features, i.e., the number of times an article is cited in various sections of the research paper.
Zhu et al. \cite{zhu} combines these approaches, presenting a list of 40 features split into five classes: count-based, similarity-based, context-based, position-based, and miscellaneous features. Among those, count-based features were the most significant. Similarly, Valenzuela et al. \cite{valenzuela} construct a set of twelve features, most of which were also similar to features generated by \cite{zhu}, and studied their importance. Pride and Knuth \cite{pride} provided their own analysis of features from \cite{zhu} and \cite{valenzuela}, narrowing down to a list of three relatively predictive features (total number of citations, abstract similarity, and author overlap) for classifying the influence of a citation. While the above methods detect the influence of a citation, Abu-Jbara et al. \cite{abu-jbara-etal-2013-purpose} explores vocabulary, number of citations in the sentence, and grammatical construction to classify the purpose of a citation. We draw upon each of these works in constructing our own features for our hand-generated features module.

More recent approaches for citation classification use deep learning techniques from natural language processing \cite{yang-etal-2016-hierarchical}.
These methods rely on word embeddings from pretrained linguistic models such as BERT \cite{bert}, ELMo \cite{elmo}, or GLoVe \cite{pennington2014glove} to transform the citation context or full-text data into vector representations. Jurgens et al. feeds these embeddings into a random forest for classification \cite{jurgens-etal-2018-measuring}, but others have used these vectors as input for neural network classifiers. 
For example, Structural Scaffolds \cite{cohan2019structural} feeds the embedded representation of the data into a bidirectional LSTM-with-attention. 
We adopt a similar approach in our LSTM-with-attention module.

\section{Task Description}\label{sec:task}
The 3C Shared Task \textit{Citation Context Classification based on Purpose} competition is a supervised multiclass classification challenge, where each citation context must be categorized based on its purpose. 
The categories (i.e., the possible labels) are:
BACKGROUND, USES, COMPARES\_CONTRASTS, MOTIVATION, EXTENSION, and FUTURE. The training dataset consists of 3000 labelled observations, 
each including the citing title, citing author, cited title, and cited author, along with the sentence in which the citation occurred. Labels, which were constructed by humans, were provided for each training example. The test dataset contains 1000 unlabelled samples, 500 of which were used for public scoring, and the other 500 of which were used for the final ranking.

\section{Methodology}\label{sec:methodology}
Our model was trained on three different tasks: classifying purpose, section, and worthiness similar to Cohan et al. \cite{cohan2019structural}.
The model consists of three modules, as shown in Figure \ref{ModelArch}, which will be discussed in this section. 

\subsection{Hand-Generated Features}


We first developed frequency-based features. To compute these, we partitioned the full text into introduction, methods, results, and discussion sections. Then we computed counts of citations in each of these sections and in the full article text, leading to five features in total.

To compute a set of relative position features, we located the sentence of the current citation and computed its relative position with respect to the number of sentences in the article. Since citations can repeat in the full paper text, we also calculated the relative position of the first appearance of the citation in the text. 


Additionally, we utilized the citation context and titles of the citing and cited articles. Specifically, for the COMPARES\_CONTRASTS label, we manually compiled a vocabulary set that is typically used in the citing context (See Appendix A for our list of key words). We then created a binary feature of whether or not a citation context contained the keywords from this vocabulary set. As our final feature, we computed the number of non-stop-words in common between the citing paper and cited paper title. Stop words were identified using Python NLTK's English stop word corpus \cite{nltk}.


\subsection{TF-IDF }
For the second module of the classifier, we generated term frequency-inverse document frequency (TF-IDF) vectors for each citation context. A TF-IDF score is a normalized count of the number of times a word appears in a given citation context relative to the remainder of the corpus. For a given citation context, a TF-IDF score is calculated as:
$$
\textrm{TF-IDF}(t,d) = \textit{tf}(t,d) \cdot \textit{idf}(t),\;\;\textrm{where}
$$
$$
\textit{idf}(t) = \log\left( \frac{1+n}{1+\textit{df}(t)}\right)+1,
$$
where \textit{tf} is the number of times a given term $t$ occurs in a document $d$, $df(t)$ is the number of documents which contain term $t$, and $n$ is the total number of documents.
To compute both term frequencies and inverse document frequencies, we used the citation context and its preceding and subsequent sentences in the full article text. 

\begin{figure}[ht!]
    \centering
    \includegraphics[width = 0.5\textwidth]{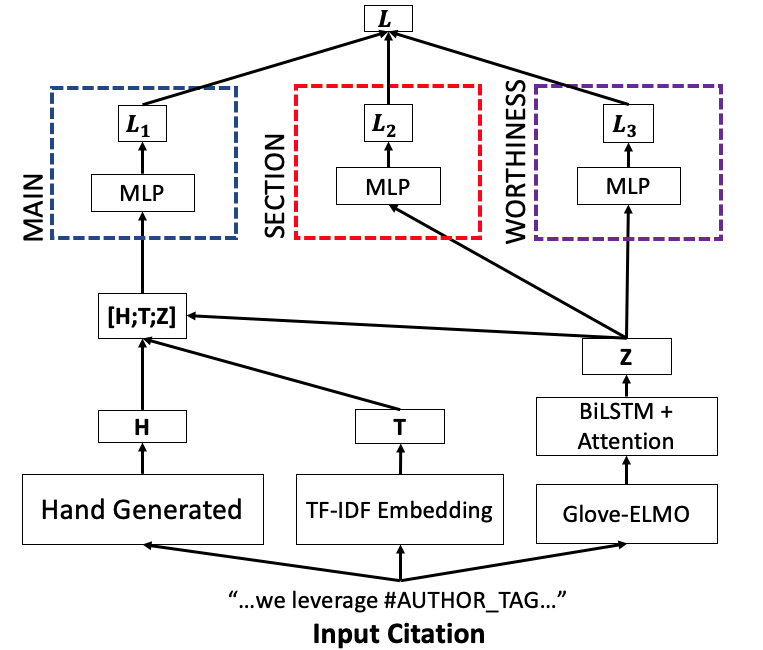}
    \caption{Model architecture with auxiliary tasks.}
    \label{ModelArch}
\end{figure}

\subsection{LSTM-with-Attention}
Our third module is based on the Structural Scaffolds model of Cohan et al. \cite{cohan2019structural}. For a given sentence, we encoded each token as a word vector. We generated the word vector by concatenating both its word representation (generated using GloVe \cite{pennington2014glove}) and contextualized embedding (generated using ELMo \cite{elmo}). Then, we passed the sequence of word vectors to a bidirectional long short-term memory network \cite{hochreiter1997long} to generate the contextual information of each token in the sentence. Finally, we fed the hidden states of the network into an attention mechanism which generates a vector encoding the input sentence.

\subsection{Module Consolidation \& Multi-Task Learning}
Following Cohan et al$.$ \cite{cohan2019structural}, we used multi-task learning to improve the performance of our model. In addition to citation-purpose classification, we included two auxiliary tasks (section and worthiness classification) that focus mostly on the structure of the article.
Since the data for these tasks are relatively easier to obtain, it could be possible to achieve better performance without 
increasing the dataset size for citation-purpose classification. The first auxiliary task is to identify citation worthiness. With a given sentence, the task is to determine whether it includes a citation. 
The second task is to predict the section title based on a given citation sentence.
For the auxiliary tasks, we utilized the auxiliary dataset generated by Cohan et al. \cite{cohan2019structural}. 
The loss of our network is a weighted sum of cross-entropy losses from each individual task. We weighed the main task (purpose classification) higher than the two auxiliary tasks. 

Our final model architecture combines all three modules into one classifier (Figure \ref{ModelArch}). The model takes the output from the LSTM attention module and concatenates it with both hand-generated and TF-IDF features before feeding it into the final multi-layer perceptron for classification. For the auxiliary tasks, only the sentence encoding from the LSTM is utilized with no additional features. 






\begin{table}[h]
    \begin{tabular}{c|c c}
    Participant & Public  & Private \\
    \hline
    IREL & 0.27968 & 0.26973 \\
    nlp\_player & 0.31385 & 0.26440 \\
    Duke Data Science & 0.25294 & \textbf{0.26325} \\
    \hline
    Our Best Test Score & 0.20825 & \textbf{0.28071}\\
    \hline
    No Hand Generated & 0.23846 & 0.22603 \\
    No LSTM + Attention & 0.17521 & 0.15470 \\
    No TF-IDF & 0.23413 & 0.21318 
    \end{tabular}
    \newline
    \caption{Model Performances. Top three competitors are shown first, then our best private score. Finally, an ablation study on our best score is shown in the bottom rows.}
    \label{leaderboard}
\end{table}

\begin{table*}[ht]
    \centering
    \begin{tabular}{c|c|c|c|c|c|c}
     & \multicolumn{6}{c} {One vs$.$ All ROC-AUC per class} \\
    Feature & BKD. & C/C & EXT. & FUT. & MOT. & USES \\
    \hline
    Num. Citations in Article & \textbf{0.424} & 0.535 & 0.516 &  \textbf{0.300} &  \textbf{0.574} &  \textbf{0.590} \\
    Vocabulary Set For COMPARES$\_$CONTRASTS & 0.489 & \textbf{0.579} & 0.501 & 0.495 & 0.486 & 0.465 \\
    Title Keyword Overlap  & 0.459 &  \textbf{0.585} & \textbf{0.662} &  \textbf{0.390} & 0.521 & 0.488 \\
    Num. Citations in Introduction & 0.531 & 0.517 & 0.472 &  \textbf{0.355} &  \textbf{0.587} &  \textbf{0.407} \\
    Num. Citations in Methods & 0.437 & 0.438 & 0.487 & \textbf{0.403} & 0.468 & \textbf{0.707} \\
    Num. Citations in Results & 0.447 & 0.534 & \textbf{0.588} & 0.433 & 0.523 & 0.531 \\
    Num. Citations in Discussion & 0.502 & 0.546 & 0.466 & \textbf{0.682} & 0.472 & 0.461 \\
    Relative Position in Full Text & 0.432 & \textbf{0.574} & 0.554 &  \textbf{0.798} &  \textbf{0.427} & 0.546 \\
    Relative Position of First Citation & 0.459 & 0.553 & 0.520 &  \textbf{0.752} & 0.434 & 0.529 \\
    \hline
    TF-IDF MLP & 0.542 & 0.518 & 0.435 & 0.294 & 0.520 & 0.651
    \end{tabular}
        \caption{Feature Metrics. Features with AUC above .57 or below .43 are in \textbf{bold}. Bottom row reports TF-IDF fed through a 2-layer MLP.}
    \label{ROCS}
\end{table*}

\section{Results} \label{sec:results}
Table \ref{leaderboard} reports our performance on the citation purpose classification task, along with scores for the top two submissions, our best private score, and results of an ablation study. Our third-place model consists of solely the LSTM-with-attention developed by \cite{cohan2019structural} using the competition dataset padded by the supplemental ACL-ARC dataset \cite{bird2008acl}. This supplemental dataset was provided by the competition organizers to help competitors tune their models with greater precision. However, our best-performing model on the private dataset was a combination of hand-generated features, TF-IDF embeddings, and an LSTM-with-attention (this model was not selected for the final leaderboard as it had lower performance on the public dataset). 


\subsection{Ablation Study} 
In order to understand the impact of each component of our best performing model, we conducted an ablation study, where we removed one module at a time to understand its individual impact on our final model, and the results are shown in the final three rows of Table \ref{leaderboard}. The LSTM-with-attention has the greatest impact on our Macro-F1 score, as when removed, performance drops the most ($\delta = -0.126$). Both TF-IDF embedding features and hand-generated features impact performance marginally, with $\delta = -0.068$ and $\delta = -0.055$ respectively. Thus, the deep learning component of our model, which represented the sequential elements of the citations, was the most important as compared to the TF-IDF embedding, which focused on vocabulary frequency in citations, and the hand-generated features, which looked at emergent syntactic or citation frequency properties.
\subsection{Feature Analysis}
To facilitate future work on this dataset, we assembled a list of the most successful features we generated by hand. Table \ref{ROCS} reports on these features and for what classes they perform the best. After generating scalar values for each predictor, we generated ROC curves in one-versus-all scenarios for each class. If the area under the curve (AUC) is greater than 0.57 or less than 0.43, we deem the feature to be a strong positive or strong negative predictor, respectively. These values are seen in bold in the table. Because the objective of this competition is to maximize F1 score, we prioritized high-signal features for smaller classes such as COMPARES$\_$CONTRASTS, FUTURE, and MOTIVATION. For example, we created a specific vocabulary set to filter COMPARES$\_$CONTRASTS citations. Most classes have 2-3 high-signal features, but BACKGROUND only has 1 due to its status as the largest class in the dataset. Features that provided signals that were too weak to be useful in this task included: Relative Position of Citation in References, Self-citation with First Author, Self-citation with Coauthors, Structural Vocabulary/Common Word Embeddings, and Number of Verbs in Sentence. Self-citation is when the citing author and cited author are the same. Note that this goes against \cite{valenzuela} and \cite{pride}, which found that self-citation was a valuable classification feature for classifying citation influence (recall that we focus on purpose classification).

Although it is impossible to generate an ROC curve based on raw TF-IDF values (because they create a matrix rather than a vector), we fed the vectors into a simple 2-layer multi-layer perceptron and reported the classifier AUC score as a proxy. These are shown in the bottom row of Table \ref{ROCS}. As we can see, even a simple model can generate strong predictions using TF-IDF data.

\section{Discussion} \label{sec:discussion}

\subsection*{Hand-generated features vs$.$ Deep Learning}
In our ablation study (Table \ref{leaderboard}), eliminating hand-generated or TF-IDF features had marginal impact on overall F1 scores, whereas eliminating the word vector embeddings generated by the LSTM with attention led to a steep drop in performance. 
In the future, substantively richer custom features could be crafted, but as we found, given the high variability of author styles and phrasing throughout the dataset, this task proves challenging. Pretrained word vector embeddings carry the advantage of being generated from massive English corpora and thus may be able to better encode the citation contexts in their wide linguistic variability. 

Deep learning approaches, however, suffer more from overfitting relative to their hand-generated peers. We found that during training, it was common for our model to easily achieve training F1 scores over 0.80 while at most achieving 0.30 on validation sets even after reducing model complexity, implementing heavy dropout, and early stopping. 



\subsection*{Author Overlap and Cross Domain Shift}

A large challenge with this task was the nature of the dataset and its train and test partition. Papers and authors in the training set are separate from papers and authors in the test set, and these papers can come from any domain of academic work. Thus, vocabulary features tended to be weak due to variance across domains, which have different norms regarding language and citations, and most importantly, domain-specific vocabulary. In addition, the training set is small enough that there is large risk of fitting to author idiosyncrasies in writing rather than broad syntactical patterns in the citations. 
A larger training set may alleviate such issues in the future.

When exploring TF-IDF vectors and simple multi-layer perceptrons, we discovered a huge boost in performance when authors and papers were allowed to overlap between train and validation sets (validation ROC-AUC scores of $\sim0.95$). Although we changed this to more accurately emulate the competition test set, resulting in the AUC scores reported in Table \ref{ROCS}, the fact that we saw good performance with a simple TF-IDF embedding illustrates the impact of author overlap on model performance. 
Future work could focus on generating features that are author and domain independent.

\subsection*{More Textual Information is Better}

We found that the full text of the citing articles was crucial for our model. Not only were our full-text hand-generated features more predictive than the ones crafted from the given citation context, our TF-IDF embeddings in the final model also utilized neighboring sentences around the citation from the full text. We often found that the given citation context was far too limited to produce meaningful features, and that the more predictive indicators were found in the preceding and subsequent sentences.
We anticipate that using the abstracts or the full texts of the \textit{cited} articles (which were not available during the competition) would further improve performance on predicting purpose, as we would be able to generate features such as abstract similarity, which Valenzuela et al. \cite{valenzuela}, and Pride and Knoth \cite{pride} found to be valuable in predicting influence. 

\bibliographystyle{acl_natbib}
\bibliography{citation.bib}

\appendix

\section*{Appendix}

\subsection*{A: Hand-generated Features}
Vocabulary set for screening COMPARES$\_$CONTRASTS: \texttt{"recently", "recent", "compared", "comparison", "similar", "studies", "reported", "others", "normally", "showed", "in line with", "despite", "relationship"}

\end{document}